\documentclass[runningheads]{llncs}
\usepackage{graphicx}
\usepackage{multirow}
\usepackage[hidelinks]{hyperref}
\usepackage{adjustbox}
\usepackage{xcolor}
\usepackage{soul}

\begin{document}
\title{Robust Burned Area Delineation through Multitask Learning
\thanks{
\small{This work was carried out in the context of the projects SAFERS (H2020, Grant ID. 869353) and OVERWATCH (HEU, Grant ID. 101082320).\\
$*$ Equal contribution.}
}}
%
%
\author{
Edoardo Arnaudo $*$\inst{1,2}\orcidID{0000-0001-9972-599X} \and
Luca Barco $*$\inst{2}\orcidID{0000-0002-9089-9616} \and
Matteo Merlo$^\ddagger$\inst{1}\orcidID{0009-0002-8008-5093} \and
Claudio Rossi\inst{2}\orcidID{0000-0001-5038-3597}
}
\authorrunning{E. Arnaudo et al.}
%
\institute{
Politecnico di Torino, Corso Duca degli Abruzzi 24, 10129 Torino, Italy
\email{name.surname@polito.it, $^\ddagger$s287576@studenti.polito.it}\\
\and
Fondazione LINKS, Via Carlo Boggio 61, 10138 Torino, Italy
\email{name.surname@linksfoundation.com}\\
}
\maketitle              
\begin{abstract}
In recent years, wildfires have posed a significant challenge due to their increasing frequency and severity. For this reason, accurate delineation of burned areas is crucial for environmental monitoring and post-fire assessment. However, traditional approaches relying on binary segmentation models often struggle to achieve robust and accurate results, especially when trained from scratch, due to limited resources and the inherent imbalance of this segmentation task. We propose to address these limitations in two ways: first, we construct an ad-hoc dataset to cope with the limited resources, combining information from Sentinel-2 feeds with Copernicus activations and other data sources. In this dataset, we provide annotations for multiple tasks, including burned area delineation and land cover segmentation. Second, we propose a multitask learning framework that incorporates land cover classification as an auxiliary task to enhance the robustness and performance of the burned area segmentation models. We compare the performance of different models, including UPerNet and SegFormer, demonstrating the effectiveness of our approach in comparison to standard binary segmentation.

\end{abstract}
\section{Introduction}
\label{sec:intro}
In recent years, wildfire events have become a recurring major problem, due to their increasing frequency and severity. 
These events have serious environmental and socio-economic impacts: with potential to cause extensive damage to forests, wildlife habitats, and even human lives. For this reason, understanding and effectively managing wildfires represents a crucial task for first responders and decision makers.

Accurate and reliable delineation of burned areas is therefore essential for various applications, including environmental monitoring and post-fire assessment.
Traditional approaches for burned area delineation often rely on binary segmentation models trained from scratch.
However, these models may struggle to achieve accurate and robust results, due to the limitations of the underlying data.
First, resources specifically tailored for this task remain particularly scarce, often lacking large and diverse datasets for an effective training.
Second, burned area segmentation is an inherently unbalanced problem, as the extent of burned areas is often significantly smaller compared to non-burned regions in the input imagery. This imbalance usually hinders the generalization abilities of the models, when applied in different scenarios.

Furthermore, existing datasets used for burned area delineation are often lacking in terms of surface covered \cite{colomba2022dataset} or diversity \cite{prabowo2022deep}. These shortcomings may hinder the ability of the models to generalize effectively, underlining the need for more comprehensive and varied data sources to enhance model performance and real-world applicability.

To address these limitations, we first construct an ad-hoc dataset, specifically tailored for the task of burned area segmentation, cross-referencing information from the Copernicus European Monitoring System (EMS) with Sentinel-2 feeds and other relevant sources. This dataset provides a comprehensive set of samples, with a focus on the European soil, including annotations for two different tasks: burned area delineation, and land cover segmentation.
Second, we propose a multitask learning framework that leverages land cover classification as an auxiliary task. By incorporating this information into the learning process, we aim to improve the robustness and performance of the model on the burned area segmentation task.
We compare the performance of different models, including UPerNet \cite{xiao2018upernet} and SegFormer \cite{xie2021segformer}, demonstrating the effectiveness of our approach against a classic binary segmentation training in several configurations, including with and without bootstrap from pretrained weights. Dataset and code related to this work are available at \href{https://github.com/links-ads/burned-area-segmentation}{github.com/links-ads/burned-area-seg}.

The remainder of this paper is structured as follows. Section \ref{sec:related} reviews related works, Section \ref{sec:dataset} describes the construction of the multitask dataset for burned area delineation, while Section \ref{sec:method} presents the proposed multitask learning framework and its components.
Section \ref{sec:experiments} details the experimental setup and discusses the obtained results. Lastly, Section \ref{sec:conclusion} concludes the manuscript, suggesting potential future directions.

\section{Related Works}
\label{sec:related}
\subsection{Aerial Semantic Segmentation}
Considering remote sensing and aerial images, semantic segmentation plays a crucial role in various applications, including urban planning \cite{mahmud2021road}, land cover monitoring \cite{arnaudo2023hierarchical}, and crisis management \cite{cambrin2022bas_transformers}.
Existing semantic segmentation methods typically rely on convolutional encoder-decoder architectures (CNNs), with different variants to capture both the global context and the finer details of the scene.
Approaches such as Fully Convolutional Networks (FCN) \cite{long2015fcn} and U-Nets \cite{ronneberger2015unet}, make use of bottleneck components to encode pixel information into semantically meaningful vectors, coupled with skip connections to integrate lower-level features.
Other solutions involve multiscale feature extraction and fusion, such as DeepLab \cite{chen2017deeplab} and PSPNet \cite{zhao2017pspnet}, where inputs are processed with varying kernel sizes and dilations to capture local and global context at once. Subsequent variants often combine these concepts to provide more robust features \cite{chen2017deeplab,xiao2018upernet}.
Segmenting aerial images introduces several specific challenges that often require tailored solutions. Unlike other settings, satellite data often provides multiple spectra beyond the visible bands.
These can be integrated in multiple ways, such as additional input channels \cite{tavera2022augmentation} or using ad-hoc encoders for feature fusion \cite{valada2017adapnet}.
Moreover, aerial images they are often denser, containing several entities against complex backgrounds, with wider spatial relationships. To address this, attention components are commonly employed to better model long-distance similarities among pixels \cite{niu2021hybrid}.
Transformer architectures and their segmentation variants \cite{xie2021segformer} thus become a natural choice in this case, given its inherent ability at extracting long-range relations.

\subsection{Burned Area Delineation}
Over the years, numerous techniques have been proposed to delineate burned areas from remote sensing data.
Standard approaches make use of several spectral indices, to discern burned soil from unburned areas using a combination of multiple bands. The \textit{de facto} standard is represented by the Normalized Burn Ratio (NBR) \cite{cocke2005nbr} and the difference NBR (dNBR) \cite{miller2007dnbr}, which are often used in combination with other indices \cite{escuin2008ndvi}. Other variants have been developed to better adapt to specific satellite feeds, such as the Burned Area Index for Sentinel-2 (BAIS) \cite{filipponi2018bais2}.
However, these approaches are usually noisy and require further manual processing to produce clean results. In some cases, such as the dNBR, a pre-wildfire image is also needed to compare the same regions before and after the event.
In the last decades, machine and deep learning techniques obtained promising results, reducing the manual effort while obtaining more robust tools. 
Supervised classification algorithms, such as Support Vector Machines (SVM) and Random Forests (RF), have been widely employed for burned area mapping \cite{ramo2017forest,knopp2020bas_cnn}. Acting on a per-pixel basis, these approaches remain effective on lower resolution feeds such as MODIS, however their lack of contextual information may result in suboptimal results on higher resolution feeds such as Sentinel-2 \cite{knopp2020bas_cnn}.
Recently, convolutional networks have been successfully employed to produce robust results on this task, especially considering post-wildfire images only. U-net segmentation architectures \cite{monaco2021bas_attention,knopp2020bas_cnn} represent the standard approach, however Transformer-based architectures have demonstrated their effectiveness in several remote sensing scenarios \cite{tavera2022augmentation}, including burned area segmentation \cite{cambrin2022bas_transformers}.   

\section{Dataset}
\label{sec:dataset}
To carry out this work, a crucial initial step involved the construction of a custom dataset specifically tailored for multitask learning, focusing on wildfire events.
Expanding on similar works in this field \cite{colomba2022dataset}, our dataset contains 171 fire events derived from Copernicus EMS\footnote{https://emergency.copernicus.eu/}. For each Area Of Interest of the event, we provide (i) the Sentinel-2 satellite imagery, (ii) a burned area annotation, derived from EMS (iii) a land cover map, derived from ESA WorldCover, and (iv) a cloud mask computed on the remote sensing input.

\subsection{Data sources}
We gather all the available large wildfire events in recent years from the catalog of the Copernicus EMS, an integral part of the Copernicus program launched by the European Union.
Within this open service, the Rapid Mapping module plays a crucial role by providing a curated set of Areas of Interest (AoI) associated with each event, where each crisis event has been carefully analyzed, and its delineation has been manually generated by a team of experts. Every AoI may provide three distinct manual annotations, named products: the First Estimate Product (FEP), the Delineation Product, and the Grading Product. The FEP consists of preliminary information about the affected territory and event, facilitating initial emergency response efforts. On the other hand, the delineation and grading products offer a more accurate and comprehensive label regarding the extent of the event and the assessment of damages.
Following the geographical coordinates and the time of the event associated with each AoI, we download the corresponding satellite images from the Sentinel-2 mission, that serve as input to the deep learning algorithms. Sentinel-2 captures data across 12 spectral bands with varying resolutions, ranging from 10 to 60 meters. In this study, we focus on the L2A product, which transforms the reflectance into Bottom-of-Atmosphere (BoA) values through atmospheric correction.

In addition to the satellite imagery and the burned area delineation maps, we also incorporate land cover data on the same area as an auxiliary target, exploiting the ESA World Cover dataset \cite{esaworldcover}. This resource offers annual maps for the years 2020 and 2021, featuring 11 distinct classes, including trees, shrubland, grassland, built-up areas, areas with sparse vegetation, water bodies and other surfaces. By integrating a more generic land cover information, we aim to enrich the semantic segmentation process with a broader understanding of the landscape dynamics, enhancing the robustness and contextual accuracy of our model.

\subsection{Data preparation}

\begin{figure}
  \centering
  \begin{minipage}[b]{0.49\linewidth}
    \centering
    \scalebox{1}[1.20]{\includegraphics[width=\linewidth]{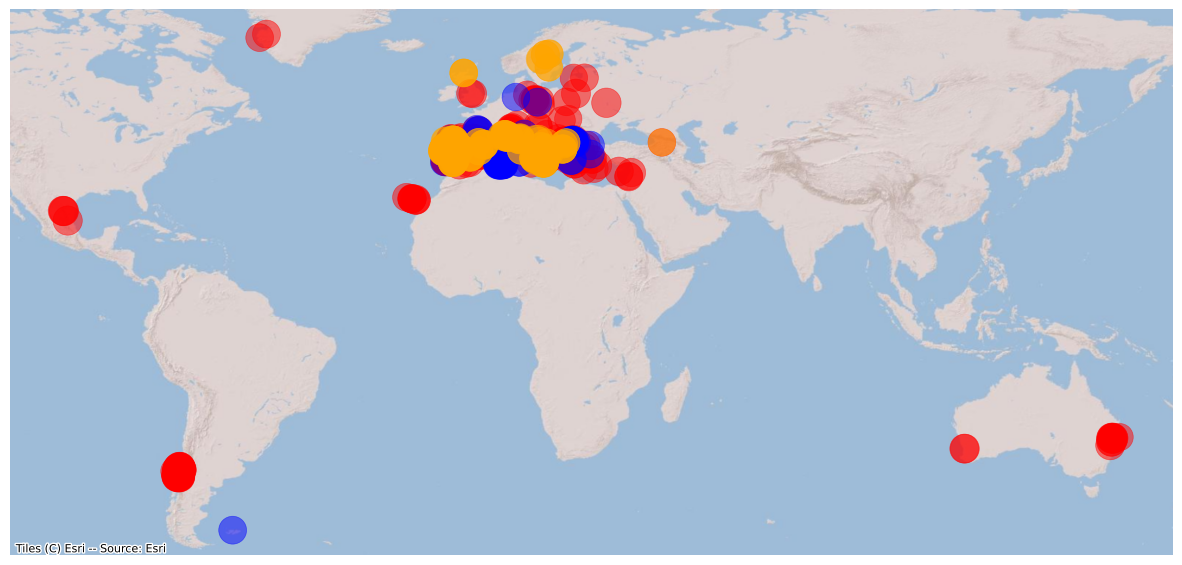}}
  \end{minipage}
  \hfill
  \begin{minipage}[b]{0.49\linewidth}
    \centering
    \includegraphics[width=\linewidth]{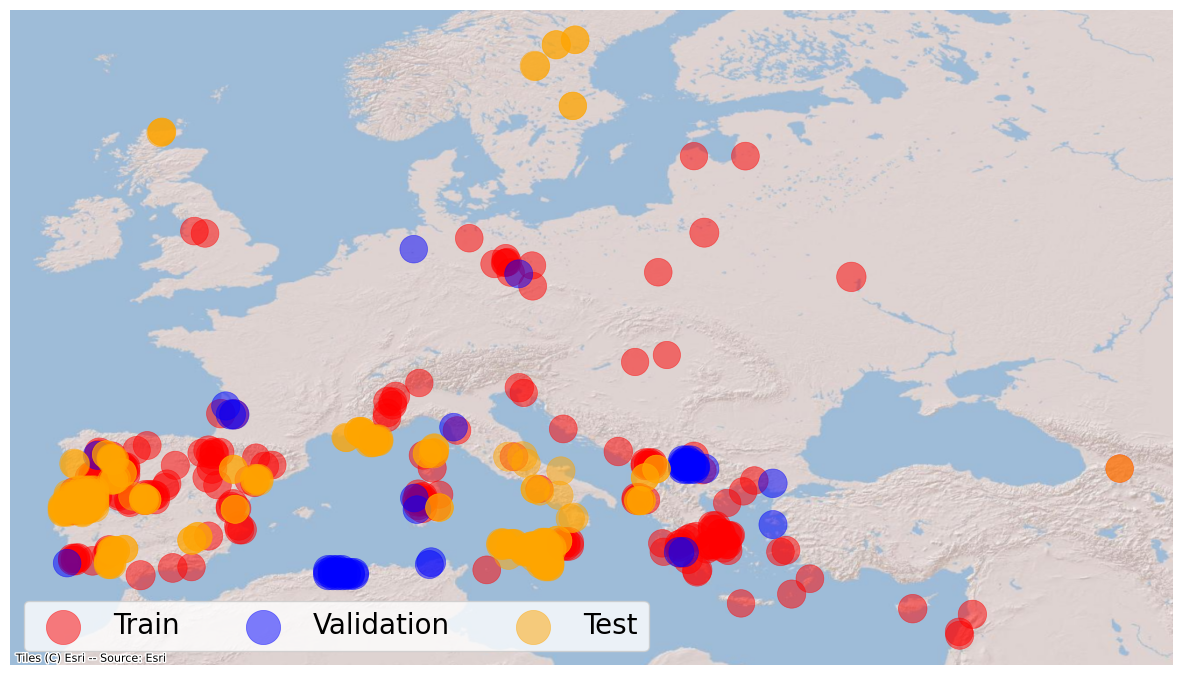}
  \end{minipage}
  \caption{Distribution of fire events contained in our dataset divided into train (red), validation (blue), and test (orange), on a worldwide scale (left) and at European level (right).}
  \label{fig:dataset-map}
\end{figure}

We download each EMS activation available, with the aim of maximizing the amount of samples with valid input image and corresponding ground truth labels. 
For each fire event we gather several details, including the event date, geographical coordinates defining the bounding box of the affected area, and corresponding delineation and grading maps. However, we note that there may be cases where the delineation, the grading, or both maps are unavailable. In such instances, given its higher quality, we maintain only those areas with a valid grading, and we generate the corresponding delineation map performing a standard binarization over the burn severity values.
Starting from the remaining processed activations, we retrieve the corresponding post-fire Sentinel-2 images, exploiting the SentinelHub services\footnote{https://www.sentinel-hub.com/}.
Given the input requirements of the models, we force each image to have a minimum dimension of 512 pixels on each side, expanding the smaller regions until this requirement is satisfied for every AoI. At the same time, we split areas larger than $2,500 \times 2,500$ pixels in multiple subsections for practical use. 
We sample and rasterize each image with a resolution of 10m per pixel, the maximum provided by Sentinel-2, upscaling the lower resolution bands with nearest neighbor interpolation.
To maximize the number of clear images, without smoke or large clouds, we consider a time frame of up to 30 days following the reported event date, selecting the satellite acquisition with the least cloud coverage.
Despite these precautions, it is not uncommon to observe clouds in the final image samples: for this reason, we further process the images using a cloud segmentation model, derived from CloudSen12 \cite{aybar2022cloudsen12}, generating a validity map. This additional mask is then applied during training, excluding every pixel covered by clouds from the loss computation. 
For the corresponding land cover maps, we retrieve the required raster layers from the ESA World Cover database, available via Microsoft Planetary Computer \footnote{https://planetarycomputer.microsoft.com/}.
No further processing is applied to the labels, except for a direct remapping from the original ESA taxonomy to a contiguous list of categories indexed from 0. A value of 255 is further assigned to pixels missing their specific category.

The final dataset comprises a collection of 433 samples, spanning from 2017 to the first months of 2023.
The events are predominantly concentrated in Europe, with select events occurring in Australia and on the American continent.
Given the same source, our collection effectively represents an extension of previous datasets \cite{colomba2022dataset}. For this reason, we dedicate every activation already present in previous works to testing purposes, training on the remaining events. This allows for easier comparisons with prior results, while also serving as a benchmark for assessing the generalizability and performance of our proposed approach.

\section{Methodology}
\label{sec:method}


\subsection{Problem statement}

The problem at hand involves developing a multitask learning framework for burned area delineation, exploiting land cover classification as an auxiliary target to guide the training. We have access to a delineation map ($y_D$) and a land cover map ($y_{LC}$) as ground truth labels. 
We employ models composed of a single encoder and a single decoder with two classification heads, namely $h_D$ and $h_{LC}$. The objective is to simultaneously train the model $f_\theta$, with parameters $\theta$, to predict accurate burned area delineations ($\hat{y}_D$), while also training on land cover classification ($\hat{y}_{LC}$) using the shared representations $\phi_{\theta}$. The shared architecture with two standard classification heads enables the model to learn from both tasks jointly.

\subsection{Framework and models}
\begin{figure}
    \centering
    \includegraphics[width=\linewidth]{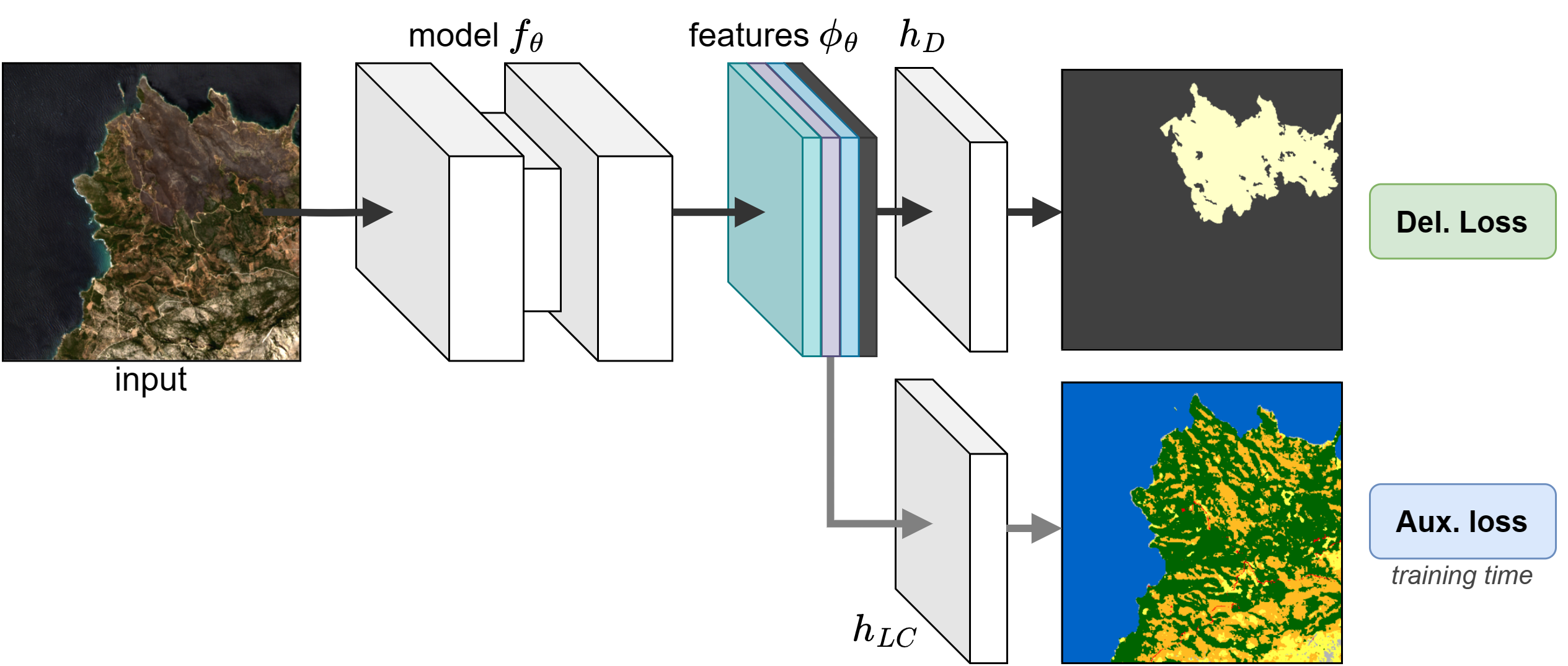}
    \caption{Multitask learning framework: the decoder features are shared with the auxiliary head $h_{LC}$ for joint traning. The auxiliary head is dropped at test time.}
    \label{fig:architecture}
\end{figure}
Our approach is shown in Fig. \ref{fig:architecture}.
To train the full model $f_\theta$ we simultaneously predict burned area delineation and land cover segmentation using the shared representations from the decoder stage $\phi_{\theta}$. These enable the model to capture and leverage common patterns and features between the two tasks, which may help in improving the segmentation outcome.
Throughout the training process, we employ a standard Cross Entropy loss, in its binary and multi-class variants respectively. The gradients derived from both tasks are jointly propagated back to update the model's parameters.
At test time, we drop the auxiliary head, focusing only on the burned area delineation performance, through standard binary segmentation.

To provide a comprehensive overview and compare standard convolutional networks with vision transformers, we explore three different architectures: two UPerNet \cite{xiao2018upernet} variants, using a Residual Network (ResNet) and a Vision Transformer (ViT) as encoders respectively, and SegFormer \cite{xie2021segformer}. 
Thanks to its unified perceptual parsing structure, UPerNet provides the flexibility to use both standard CNNs, and recent transformer-based solutions. This allows for a better comparison between the two architectures.
On the other hand, SegFormer represents an alternative end-to-end solution which demonstrated its effectiveness on aerial tasks, including burned area delineation \cite{cambrin2022bas_transformers,tavera2022augmentation}.

\section{Experiments}
\label{sec:experiments}

\subsection{Implementation details}
As mentioned in Section \ref{sec:dataset}, we train our models on the subset of activations that are not present in previous datasets \cite{colomba2022dataset}, considering the remaining ones as our test set.
We further extract a 10\% of activations from our training for validation purposes, obtaining a total of 129 wilfdire events in training, 15 in validation, and 27 for testing purposes.
To cope with the varying image dimensions, we implement a random sampling strategy that extracts square crops of $512 \times 512$ pixels from random image sections at runtime during training. For validation and testing, we adopt instead a sequential sampling strategy with overlapping tiles, reconstructing the original inputs by means of a smooth blending using splines.
We consider two groups of experiments: first, we only focus on burned area delineation, as a single training task. Second, we conduct a multitask training, using both delineation and land cover maps. In the latter case, we further mask out the burned pixels from the annotation, to avoid inconsistent labels.
For both scenarios, we train three architectures: UperNet with two different encoders (i.e., ResNet-50 and ViT-S) and SegFormer with MiT-B3 as encoder.
Moreover, we investigate the impact of using pretrained weights on the backbones in both configurations. We exploit pretrained weights derived from large-scale pretraining on SSL4EO-S12 \cite{wang2023ssl4eos12} for ResNet and ViT, in the RN50 and ViT-S variants, while we adopt weights pretrained on ImageNet \cite{xie2021segformer} for SegFormer for lack of better options.
We base our code on the \textit{mmsegmentation}\footnote{https://github.com/open-mmlab/mmsegmentation} library, adapting the model inputs to adjust for the additional channels.
In every experiment, we train on a single NVIDIA A100 GPU for 30 epochs, using a batch size of 32 tiles, AdamW as optimizer with a learning rate of 1e-4, and a Cross Entropy loss for both tasks, in binary and multi-class versions respectively. 
Following similar works \cite{cambrin2022bas_transformers,tavera2022augmentation}, we adopt macro-averaged F1 score and Intersection over Union (IoU) as evaluation metrics in every configuration.

\subsection{Results}

\begin{table}[ht]
\adjustbox{max width=\textwidth}{
\begin{tabular}{ccllcc}
\textbf{}                                         & \textbf{}                                        & \multicolumn{2}{c}{\textbf{From scratch}}                                                                                                            & \multicolumn{2}{c}{\textbf{Pretrained}}                                                              \\ \hline
\multicolumn{1}{|c}{\textbf{Setting}}             & \multicolumn{1}{c|}{\textbf{Model}}              & \multicolumn{1}{c}{\textbf{F1}}                                & \multicolumn{1}{c|}{\textbf{IoU}}                                                   & \textbf{F1}                            & \multicolumn{1}{c|}{\textbf{IoU}}                           \\ \hline
\multicolumn{1}{|c}{}                             & \multicolumn{1}{c|}{\textbf{SegFormer (MiT-B3)}} &  \textbf{89.01$\pm$ 1.39} & \multicolumn{1}{l|}{ \textbf{80.22$\pm$ 2.25}} & 90.79$\pm$ 0.46          & \multicolumn{1}{c|}{83.13$\pm$ 0.78}          \\
\multicolumn{1}{|c}{}                             & \multicolumn{1}{c|}{\textbf{UPerNet (RN50)}}     &  82.33$\pm$ 9.17          & \multicolumn{1}{l|}{ 70.94$\pm$ 12.63}         & \textbf{91.27$\pm$ 0.08} & \multicolumn{1}{c|}{\textbf{83.95$\pm$ 0.13}} \\
\multicolumn{1}{|c}{\multirow{-3}{*}{\textbf{STL}}} & \multicolumn{1}{c|}{\textbf{UPerNet (ViT-S)}}    &  87.65$\pm$ 2.01          & \multicolumn{1}{l|}{ 78.08$\pm$ 3.17}          & 89.20$\pm$ 1.29          & \multicolumn{1}{c|}{80.53$\pm$ 2.09}          \\ \hline
\multicolumn{1}{|c}{}                             & \multicolumn{1}{c|}{\textbf{SegFormer (MiT-B3)}} &  \textbf{90.94$\pm$ 0.17} & \multicolumn{1}{l|}{ \textbf{83.38$\pm$ 0.29}} & 90.91$\pm$ 0.28          & \multicolumn{1}{c|}{83.34$\pm$ 0.47}          \\
\multicolumn{1}{|c}{}                             & \multicolumn{1}{c|}{\textbf{UPerNet (RN50)}}     &  89.82$\pm$ 1.76          & \multicolumn{1}{l|}{ 81.57$\pm$ 2.87}          & \textbf{91.86$\pm$ 0.30} & \multicolumn{1}{c|}{\textbf{84.94$\pm$ 0.51}} \\
\multicolumn{1}{|c}{\multirow{-3}{*}{\textbf{MTL}}} & \multicolumn{1}{c|}{\textbf{UPerNet (ViT-S)}}    &  89.76$\pm$ 0.15          & \multicolumn{1}{l|}{ 81.43$\pm$ 0.25}          & 90.69$\pm$ 0.58          & \multicolumn{1}{c|}{82.98$\pm$ 0.97}          \\ \hline
\end{tabular}
}
\label{tab:res}
\caption{Experimental results in single (STL) and multitask (MTL) training, comparing models trained from scratch, or using pretrained encoders.}
\end{table}

\begin{table}[]
\centering
\adjustbox{max width=\textwidth}{
\begin{tabular}{|cc|c|c|}
\hline
\textbf{Setting}            & \textbf{Model}            &  \multicolumn{1}{c|}{\textbf{\begin{tabular}[c]{@{}c@{}}Training time (1 Ep.)\end{tabular}}} &  \multicolumn{1}{c|}{\textbf{Param. (M)}} \\ \hline
\multirow{3}{*}{\textbf{STL}} & \textbf{SegFormer (MiT-B3)} & 3h28m                                         & 44,6                                 \\
                            & \textbf{UPerNet (RN50)}     & 3h20m                                           & 64,1                                \\
                            & \textbf{UPerNet (ViT-S)}    & 3h20m                                           & 57,9                                \\ \hline
\multirow{3}{*}{\textbf{MTL}} & \textbf{SegFormer (MiT-B3)} & 3h40m (+12m)                                      & 44,6                                 \\
                            & \textbf{UPerNet (RN50)}     & 3h50m  (+30m)                                          & 64,1                                \\
                            & \textbf{UPerNet (ViT-S)}    & 3h30m (+10m)                                          & 57,9                                \\ \hline
\end{tabular}
}
\label{tab:comparison}
\caption{Analysis of the computational costs in terms of training time over one epoch, as average of three epochs, and total network parameters. While the training time increases by a small margin, the parameter increase is effectively negligible given the shared encoder-decoder structure.}
\end{table}

\begin{figure}
  \centering

  \includegraphics[width=1\textwidth]{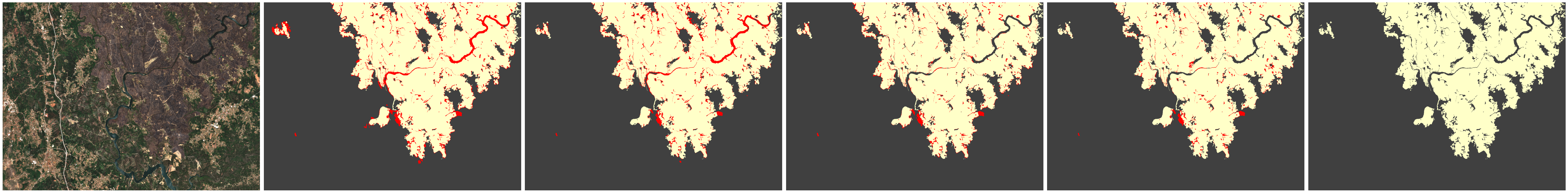}
  \hfill
  \includegraphics[width=1\textwidth]{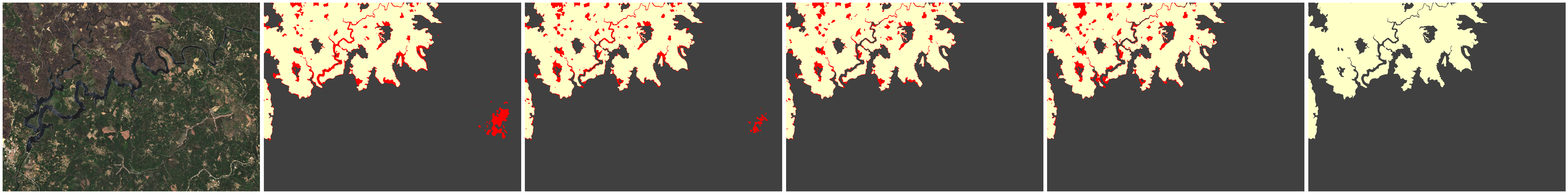}
  \hfill
  \includegraphics[width=1\textwidth]{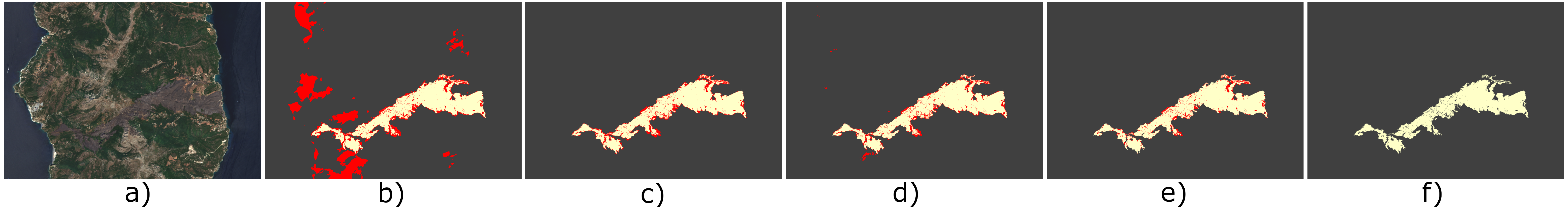}

  \caption{Qualitative examples derived from UPerNet-RN50: a) Sentinel-2 input; b) Single task and c) Multitask from scratch; d) Single Task and e) Multitask with pretrained weights; f) ground truth. Red pixels represent prediction errors.}
  \label{fig:qualitative}
\end{figure}

We conduct single-task (STL) and multitask (MTL) training experiments using both pretrained and non-pretrained weights. For each configuration, we perform three separate runs with different seeds, reporting the results in Table \ref{tab:res} as average scores with their corresponding standard deviation.
Focusing on the experiments conducted without pretrained weights, the multitask setting consistently achieves superior performance and lower standard deviation compared to the single task setting.
Except for the SegFormer, that reports the highest scores in both variants, the multitask approach exhibits a noticeable improvement of $+3.85$ in terms of F1 score, or $+5.71$ in terms IoU, averaged across every model. Furthermore, we note that the results are way more stable in the multitask configuration, where the standard deviation decreases by $-3.51$ (F1) and $-4.88$ (IoU). This is also shown in Figure \ref{fig:qualitative}, where the latter produce more reliable segmentation maps.
Considering the experiments with pretrained weights, the disparity between single and multitask performance is no longer apparent, with higher and more stable scores even in the single task setup. This is expected, as large-scale pretraining has been proven to be effective in several contexts \cite{wang2023ssl4eos12}.
Nevertheless, multitask training still yields an average overall improvement of $+0.73$ in F1 score and $+1.21$ in IoU, regardless of the underlying architecture.
Lastly, comparing training from scratch and using pretrained weights, the latter always exhibit higher performances, even more so in multitask configuration. Specifically, when comparing the top performing models from both tables (i.e., Segformer in the first case, UPerNet-RN50 in the second case), the single task setting achieves $+2.24$ in F1 score and $+3.72$ in IoU, whereas in multitask achieves $+0.92$ in F1 score and $+1.56$ in IoU.
Overall, the results demonstrate the validity of the multitask strategy, exhibiting increased performance robustness, comparable to or even surpassing pretrained solutions in certain instances. 

In Table \ref{tab:comparison} we also compare the computational costs of the STL approach compared to the MTL solution. We observe that the MTL versions, despite including an additional segmentation head, exhibit only a modest increase in training speed compared to their single task learning (STL) counterparts, with a marginal difference of 20 seconds in training speed. Moreover, while MTL models do incur a slight increase in memory usage, this increment remains negligible and does not substantially impact the feasibility of implementation. This is expected, since the MTL setting only effectively adds the parameters of a single pixel classification head, which boils down to a $1 \times 1$ convolutional layer with $|\phi_\theta|$ feature channels as input, and 11 categories as output.  
Moreover, we note that during inference the auxiliary head is omitted, effectively eliminating any computational overhead associated with the auxiliary task.

\section{Conclusion}
\label{sec:conclusion}
In this work, we propose a multitask approach for burned area delineation, exploiting land cover classification as an auxiliary target. Results show that the devised solutions yield more stable and robust performances, comparable to or even superior to pretrained solutions.

Multitask learning offers promising results, especially in absence of pretrained solutions.
Despite the robust performance, the current multitask approach presents some limitations: first, the performance of the models heavily rely on the quality of the annotations of both tasks. Second, the improved robustness comes at the cost of additional computational complexity, which may limit the scalability.
Moreover, we recognize the need to delve deeper into the impact of task characteristics and explore a wider array of auxiliary tasks for a more comprehensive multi-task learning approach, including for instance multiple training objectives to further enhance scalability and generalization capabilities of the models.


Future studies may therefore focus on improving the multitask capabilities by integrating multiple heterogeneous tasks at the same time \cite{bastani2022satlas}, or may consider more computationally demanding approaches such as large-scale self-supervised learning, to generate better pretrained solutions and thus translate these downstream tasks in simpler and faster fine-tuning objectives.

%
%
%
\bibliographystyle{splncs04}
\bibliography{references}
\end{document}